\documentclass[runningheads]{llncs}
\usepackage{graphicx}
\usepackage{amsmath}

\usepackage{xcolor}

\begin{document}
\title{SuperNCN: Neighbourhood consensus network for robust outdoor scenes matching}
\titlerunning{SuperNCN}
%
\author{Grzegorz Kurzejamski\inst{1}\orcidID{0000-0002-2918-497X} \and
Jacek Komorowski\inst{1}\orcidID{000-0001-6906-4318} \and
Lukasz Dabala\inst{1}\orcidID{0000-0003-1324-7224} \and
Konrad Czarnota\inst{1} \and
Simon Lynen\inst{2}\orcidID{0000-0002-6421-541X} \and
Tomasz Trzcinski\inst{1}\orcidID{0000-0002-1486-8906}}
\authorrunning{G. Kurzejamski et al.}
%
\institute{Institute of Computer Science, Warsaw University of Technology, Poland
\and
Google
}
\maketitle              
\begin{abstract}
In this paper, we present a framework for computing dense keypoint correspondences between images under strong scene  appearance changes.
Traditional methods, based on nearest neighbour search in the feature descriptor space, perform poorly when environmental conditions vary, e.g. when images are taken at different times of the day or seasons.
Our method improves finding keypoint correspondences in such difficult conditions.
First, we use Neighbourhood Consensus Networks to build spatially consistent matching grid between two images at a coarse scale. 
Then, we apply Superpoint-like corner detector to achieve pixel-level accuracy. 
Both parts use features learned with domain adaptation to increase robustness against strong scene appearance variations. 
The framework has been tested on a RobotCar Seasons dataset, proving large improvement on pose estimation task under challenging environmental conditions.
\end{abstract}

\section{Introduction}
\label{sec:intro}
Finding correspondence between keypoints detected in images is a key step in many computer vision tasks, such as panorama stitching, visual localization or camera pose estimation.
Traditional approaches rely on detecting keypoints in images, computing their descriptors and nearest neighbour search in the feature descriptor space. 
Many works aim at making keypoint detection and description methods robust to changes of viewing conditions. \cite{SIFT1,AKAZE} investigated keypoint matching capabilities under global affine geometric transformations and image degradation by modelled noise. 
In practical applications, scene appearance can change significantly due to different atmospheric conditions, seasons or time of the day. An example of image degradation caused by these factors is shown in Figure~\ref{fig:RCS_examples}.
Scene appearance changes caused by these factors are complex, non-linear and difficult to model analytically.



Existing camera pose estimation approaches are commonly based on finding correspondences between local features detected in images. First, putative matches are found by using nearest neighbour search in the feature descriptor space. Then, matches are filtered using a robust model parameter estimation method, such as RANSAC~\cite{Fischler:1981:RSC:358669.358692}, with an appropriate geometric consistency criteria, such as epipolar constraint.

\begin{figure}
\centering
\begin{tabular}{cc}
\includegraphics[width=0.49\textwidth]{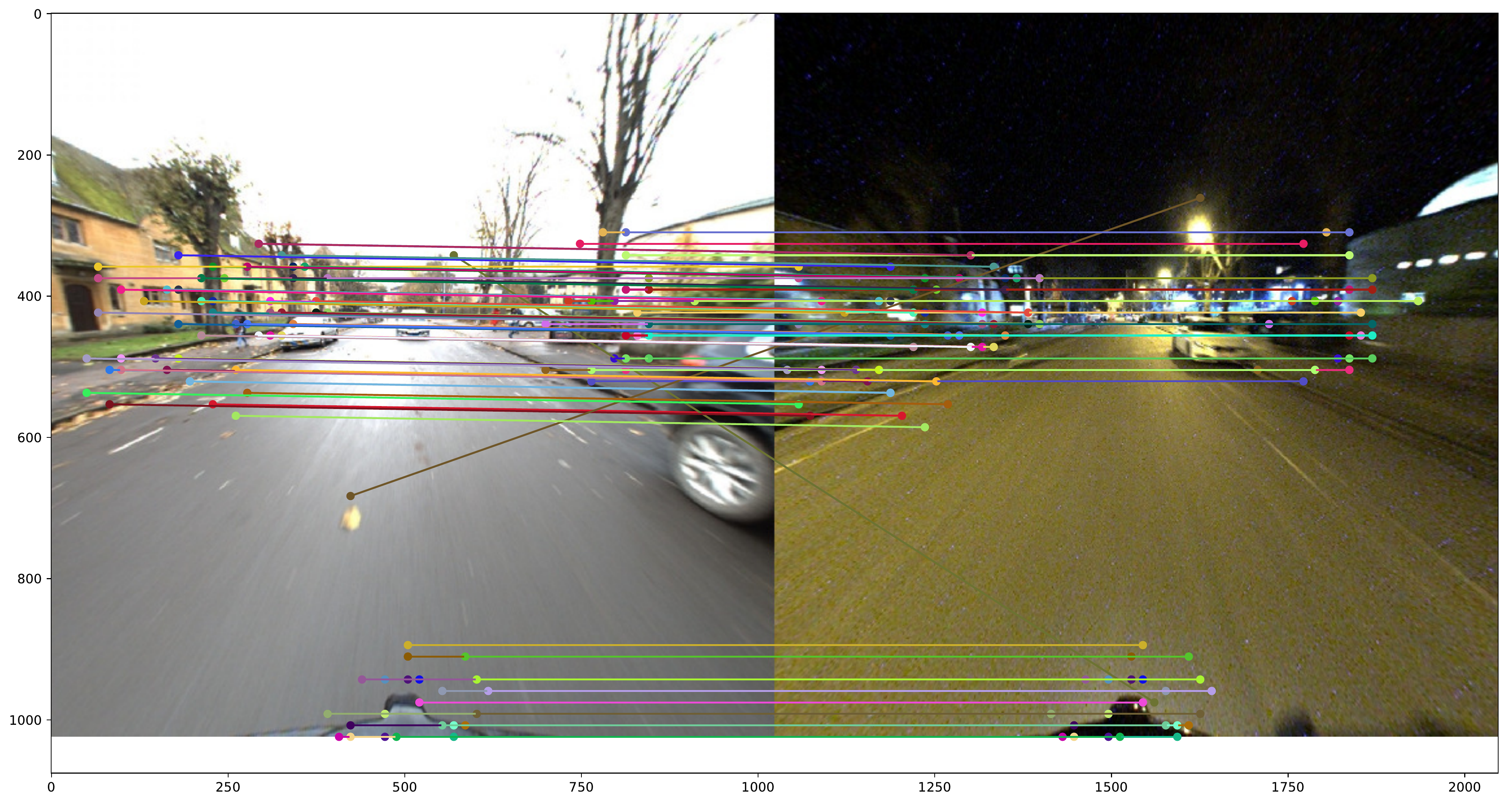}&
\includegraphics[width=0.49\textwidth]{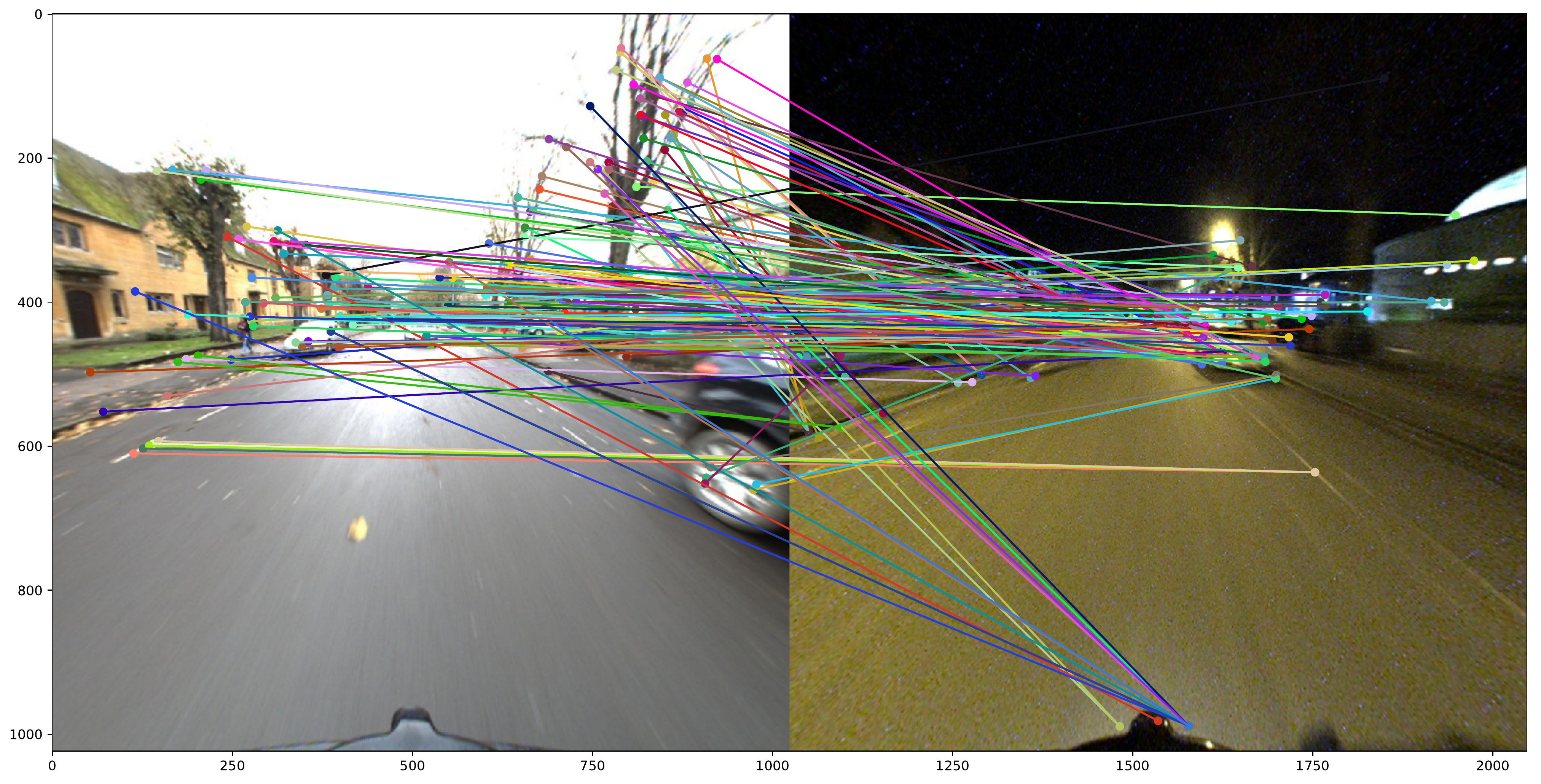}\\
(a) Our algorithm & (b) SIFT + KNN \\
\includegraphics[width=0.49\textwidth]{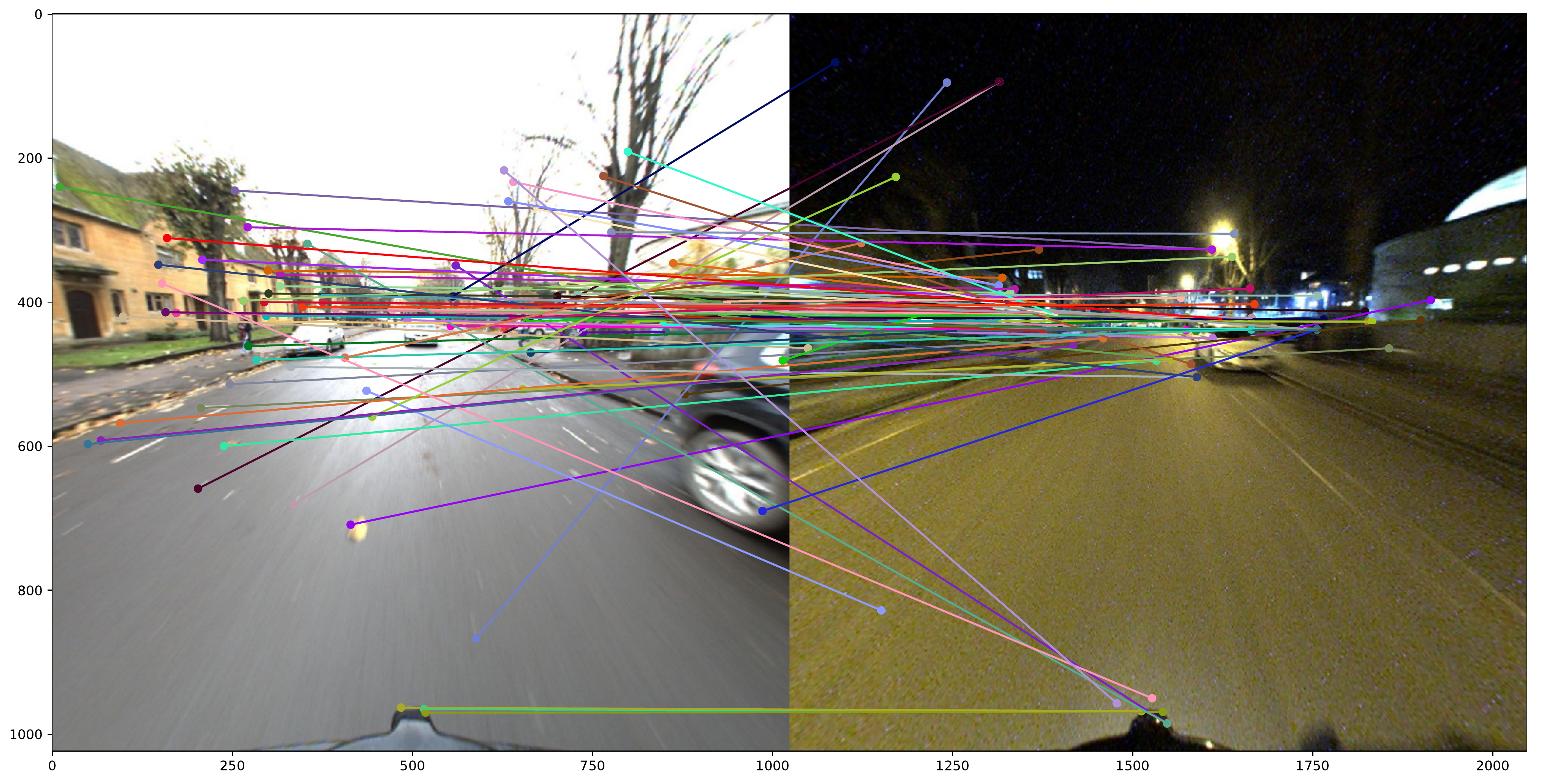}&
\includegraphics[width=0.49\textwidth]{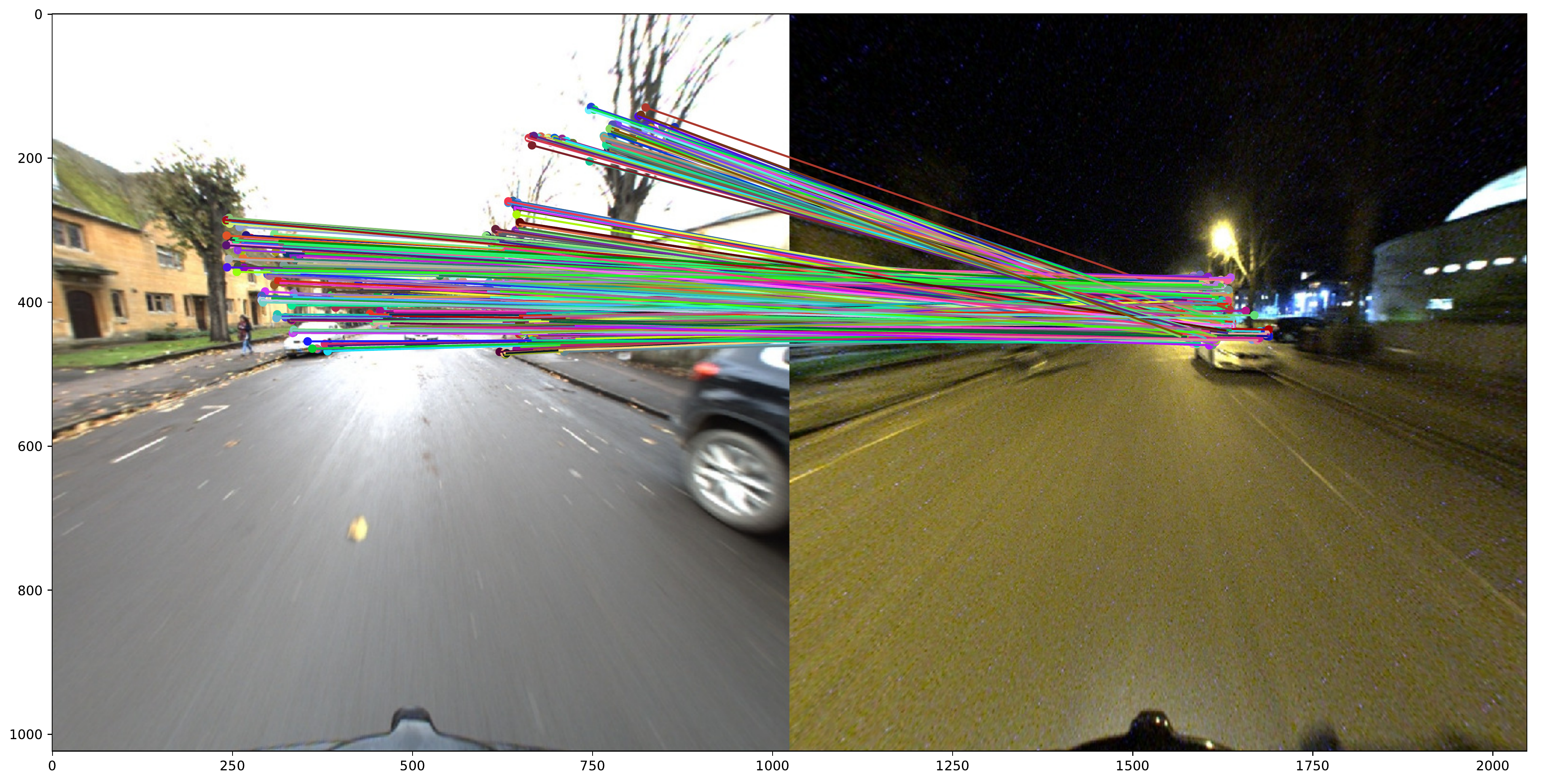}\\
(c) Superpoint + KNN & (d) ORB + GMS
\end{tabular}
\caption{Results of different feature descriptor matching strategies. 
Our method yields consistent correspondences despite strong scene appearance change.
For brevity, We show only a random subset of all three thousand correspondences found by our method.}
\label{fig:RCS_match_example}
\end{figure}

In this paper, we propose a feature matching method robust to significant changes of scene appearance.
The method combines correspondence estimation at a coarse scale, based on Neighbourhood Consensus Networks (NCNet)~\cite{NCNet}, with pixel-level accuracy, precise keypoint detection.
By using domain adaptation technique~\cite{UDAB}, the proposed algorithm is robust to scene appearance changes caused by time of the day or seasonal differences.
Unlike traditional feature matching approaches, our method imposes semi-global spatial consistency of established correspondences and integrates matching and correspondence filtering steps. This boost the performance under extreme scene appearance variations.
See Fig.~\ref{fig:RCS_match_example} for sample results.


Matches produced by NCNet are dense, spatially consistent but are at a coarse scale and have a limited spatial resolution. Each location in a feature map processed by NCNet corresponds to the block of 16x16 pixels. To allow an accurate camera pose estimation, we further push the accuracy to pixel-level.
This is achieved by detecting keypoints in original resolution images and matching keypoints in each pair of grid cells matched at a course scale by NCNet.
Architecture of our keypoint detector is similar to work of DeTone~\textit{et al}.~\cite{Superpoint}.
The resultant correspondences are spatially consistent at a course scale and have pixel-level accuracy required for precise camera pose estimation.


We evaluate our method on RobotCar Seasons dataset using different traversals in different weather and daytime conditions and compare its performance against state of the art keypoint matching techniques. Our method shows significantly better results under heavy scene appearance changes than traditional techniques based on feature matching using nearest neighbour search.







\section{Related Work}
\label{sec:related_work}
From the very beginning of SLAM~\cite{ekfslam1996}, common methods for data association are based on nearest neighbour search in a feature descriptor space.
The traditional approach is to use hand-crafted feature detectors and descriptors, such as SIFT~\cite{SIFT1}, ORB~\cite{ORB} or AKAZE~\cite{AKAZE}.
Recently, neural-network based feature detectors and descriptors, such as Superpoint~\cite{Superpoint}, gain popularity.
However, matching sparse features is challenging under strong viewpoint or illumination changes due to large scene appearance changes.

Through the years, the data association step was a bit overlooked in research, which focused more on the improvement of feature detection and descriptors.
There are only few works proposing an efficient feature matching method alternative to nearest neighbour search.
Recent advances include local neighbourhood consensus based  algorithms~\cite{DBLP:conf/cvpr/BianLMYNC17} to add robust filtering stage after nearest neighbours step.
In~\cite{trzcinski2018scone} authors propose a discriminative descriptor that includes not only local feature representation, but also information about the geometric layout of neighbouring keypoints.
There's a number of works related to optical flow and using deep neural nets to estimate the optical flow between a pair of images ~\cite{weinzaepfel:hal-00873592,DBLP:journals/corr/RanjanB16}. The optical flow methods give good results for light geometric transformations, but they are not suited for outdoor images, taken at various times of a day and from significantly different viewpoints. 
Interesting idea of determining dense correspondences between images with strong viewpoint, using hierarchical correlation architecture, is introduced in~\cite{deepmatching2016}. 
Recently, \cite{NCNet} proposes Neighbourhood Consensus Networks (NCNet), a robust method for establishing dense correspondences between two images. The method enforces semi-global spatial consistency between correspondences, and is robust to significant scene appearance changes. However its practical usage is limited to relatively low resolution images, due to usage of computationally-demanding 4D convolution.

One of the key challenges in computer vision is to devise features invariant to scene appearance changes. Large changes in illumination and viewpoint or changes due to seasonal or time-of-the-day variations have a strong impact on scene appearance. Feature descriptors of the same scene captured under different conditions can be significantly different. This adversely impacts performance of feature correspondence estimation process. 
To increase robustness against scene appearance changes due to seasonal or time-of-the-day differences, we use domain adaptation technique proposed in~\cite{UDAB}. 

\section{Network architecture}
\label{sec:architecture}
\begin{figure*}
\centering
\includegraphics[width=0.9\textwidth]{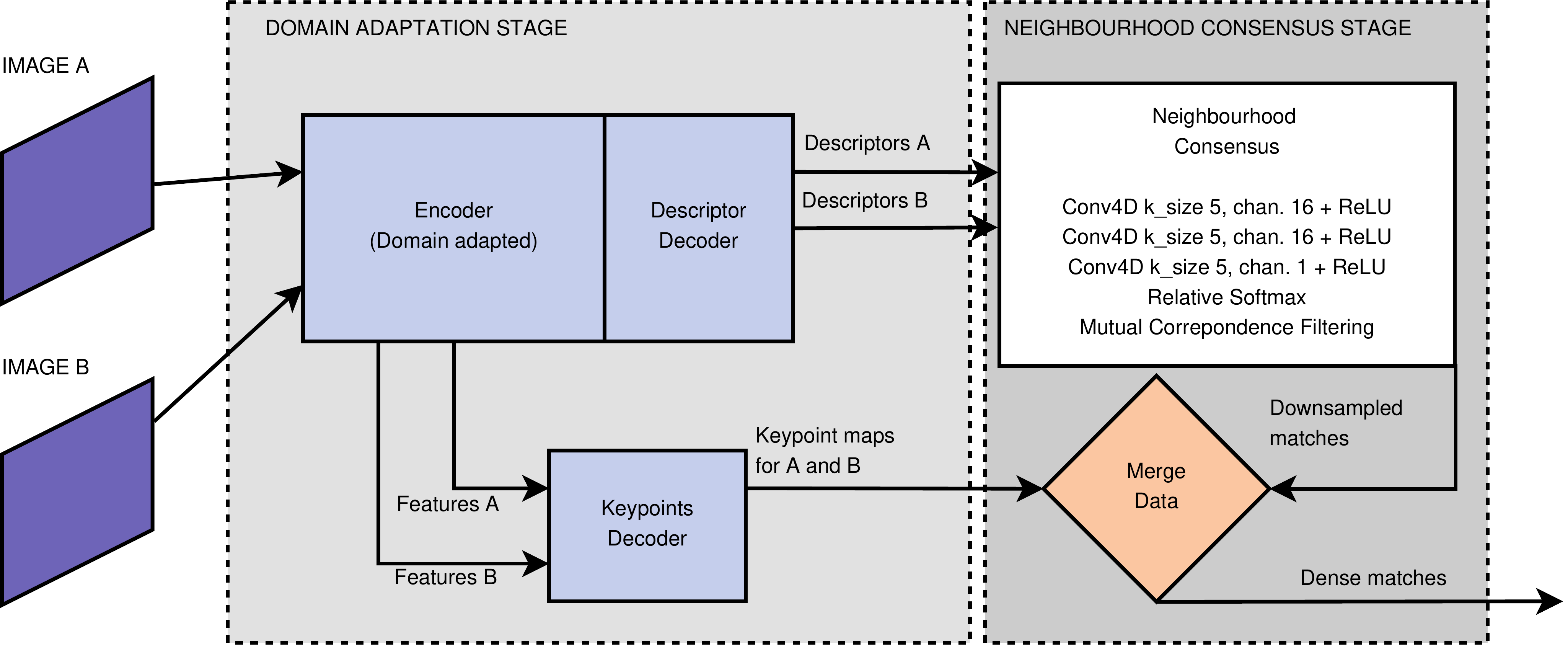}
\caption{Architecture overview of our solution.}
\label{fig:our_net}
\end{figure*}

High-level architecture of our solution is presented in Figure \ref{fig:our_net}. 
It consists of two parts: Domain Adaptation and Neighbourhood Consensus stage, trained independently from each other.
The first part is based on Superpoint~\cite{Superpoint} keypoint detector and descriptor architecture.
Encoder is a VGG-style feature extraction network trained with domain adaptation using adversarial loss~\cite{UDAB}.
It produces a domain-adapted convolutional feature map with spatial resolution decreased by 16 compared to the input image size. The feature map is fed into a Descriptor Decoder network computing dense feature descriptor map. 
The feature map produced by Encoder is forked into Keypoint Decoder. It uses sub-pixel convolution~\cite{shi2016real} (pixel shuffle) to create full resolution keypoint response map, indicating precise keypoint locations. 
Second part, Neighbourhood Consensus stage, is based on Neighbourhood Consensus Networks~\cite{NCNet} architecture. 
It takes as an input two dense feature descriptor maps computed from input images.
These maps have reduced spatial resolution, where each spatial location in the feature map corresponds to 16x16 pixel block in the input image. 
This allows efficient computation of dense correspondences between feature maps taking into account spatial consistency between all matches.
Then, to achieve pixel-level accuracy, in each 16x16 pixels block we find four keypoints by taking argmax in the corresponding region of the keypoint response map. The keypoints are matched between corresponding blocks using nearest neighbour search in the descriptor space.
For each spatial location in the feature map (corresponding to block of 16x16 pixels) we find four keypoint matches. This is depicted as Merge Data block in Figure \ref{fig:our_net}.
The resultant correspondences are spatially consistent at a course scale and have pixel-level
accuracy needed for camera pose estimation.

\subsection{Domain adaptation stage}
The first stage of our solution is based on Superpoint~\cite{Superpoint} architecture.
It uses convolutional neural network for detection and description of corner-like keypoints. It employs non-maximum suppression to enforce even distribution of detected keypoints across the image.
In our preliminary experiments, we found that using grid-based detectors, with even distribution of keypoints, improves the results of camera pose estimation. 
Traditional keypoint detectors, such as SIFT or AKAZE, tend to cluster detected keypoints in some parts of the image, which has adverse effect on accuracy of the pose estimation task.

Encoder uses a VGG-style architecure to compute the feature map from the input image. 
It consists of convolutional layers, spatial downsampling via pooling and non-linear ReLU activation functions computing feature maps with descreased spatial resolution and increased number of channels.
It uses four max-pooling layers with a stride two, each decreasing the spatial dimensional of the processed data by two. The output from the third max-pooling layer, with the spatial dimensions $W/8$ x $H/8$ and 128 channels, where W, H is size of the input image, is forked to the Keypoint Decoder head. Keypoint Decoder consists of two convolutional layers with ReLU non-linearity and produces a 65-channel tensor with $W/8$ x $H/8$ spatial resolution. The first 64 channels correspond to non-overlapping 8x8 pixel regions and the last channel channel corresponds to 'no keypoint' dustbin. Thus, each spatial location of the resulting $W/8$ x $H/8$ feature map corresponds to 8x8 pixel region and values at different channels correspond to the probability of a keypoint location at each pixel of this region (with the last channel corresponding to the probability of no keypoints in the region). The output from the forth max-pooling layer, with the spatial dimensions $W/16$ x $H/16$ and 256 channels, is fed to Descriptor Decoder head. Descriptor Decoder consists of two convolutional layers with ReLU non-linearity and produces a 256-channel descriptor map with $W/16$ x $H/16$ spatial resolution. Each spatial location of the descriptor map corresponds to 16x16 pixel region of the input image.


\subsection{Neighbourhood consensus stage}
Neighbourhood Consensus module is based on NCNet~\cite{NCNet} method. 
First it computes four dimensional similarity map between all possible pairs of locations (each being two dimensional point) in two input feature maps. 
Then, four-dimensional convolutions are applied to modify score of each 4D similarity map location depending on the score of neighbourhood locations. This enforces spatial consistency between established correspondences. Soft mutual neighbourhood filtering is applied, to boost the score of reciprocal matches.
Finally, for each location in one feature map, a single matching location in the other feature map is selected by choosing the location with maximal similarity score.
Due to application of 4D convolutions, matching score between two 2D locations is influenced by matching scores of neighbourhood locations. This results in more spatially consistent matches compared to the simple feature matching approach based on a nearest neighbour search in the descriptor space.
Because our method operates on downscaled feature maps, with resolution reduced by 16 in each dimension, using four dimensional convolution on a Cartesian product of 2D locations in each feature map is is computationally feasible.


\section{Training}
\label{sec:training}
This section describes the training process for our solution. We use  RobotCar-Seasons~\cite{OxfordSeasons} dataset, described in section \ref{sec:database}. 
Domain-adapted Encoder, Descriptor Decoder and Keypoints Decoder Detector are trained independently from Neighbourhood Consensus stage. First is described in section \ref{sec:sp_training}, latter in section \ref{sec:nc_training}.

\subsection{Database}
\label{sec:database}
RobotCar-Seasons\cite{OxfordSeasons} is a subset of Oxford Robotcar Dataset\cite{robotcar_dataset}. It contains sequences of images captured during multiple traversals through the same route in different weather or season conditions (dawn, dusk, night, night-rain, overcast-summer, overcast-winter, rain, snow, sun). 
Each traversal contains images from three cameras: left, right and rear.
One sequence (overcast-reference), captured at favourable conditions, is marked as a reference traversal. The ground truth contains an absolute pose of each image with respect to the world coordinate frame and intrinsic parameters of each camera.

\begin{figure}
\centering
\includegraphics[width=0.30\textwidth]{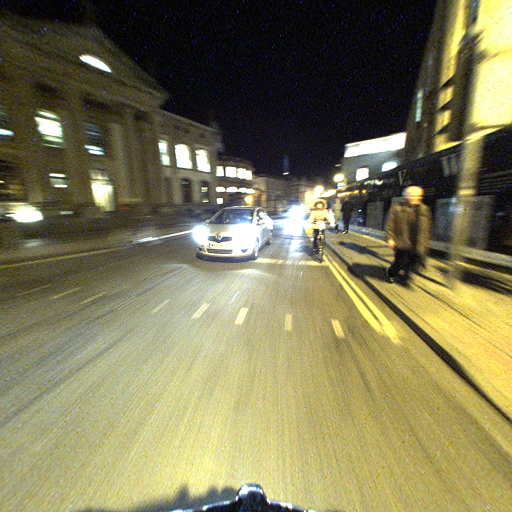}
\includegraphics[width=0.30\textwidth]{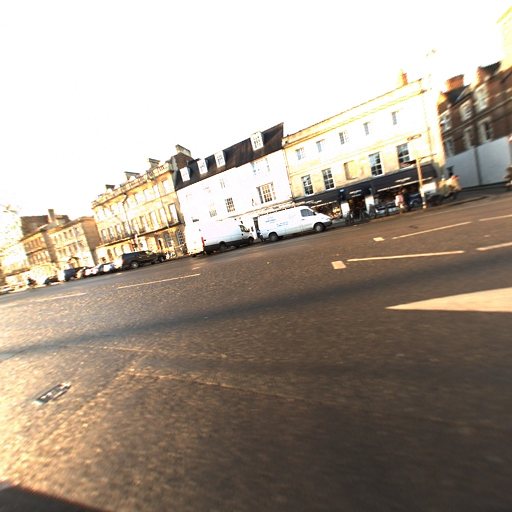}
\includegraphics[width=0.30\textwidth]{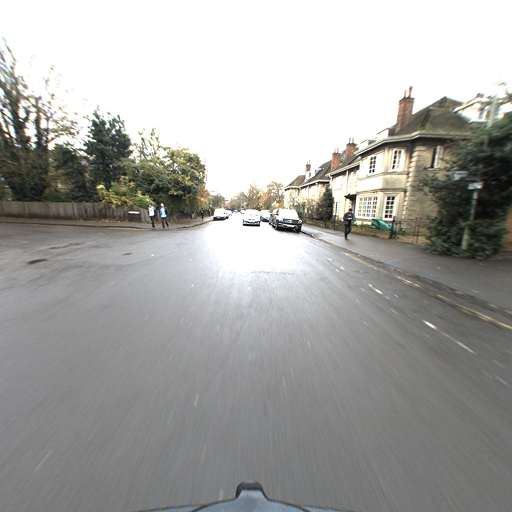}

\caption{Examples from RobotCar Seasons dataset. From the left: night, dusk and overcast samples. We can see blurring, overexposition and low texture content in various images.}
\label{fig:RCS_examples}
\end{figure}

Images have a relatively low resolution and contain all sorts of artifacts present in images acquired by real-world outdoor camera systems. 
There are overexposured areas, significant blur, light flares and compression artifacts in many images.
Exemplary images are shown in Figure \ref{fig:RCS_examples}. 

\subsection{Domain adapted feature descriptor and detector}
\label{sec:sp_training}

\begin{figure*}
\centering
\includegraphics[width=0.7\textwidth]{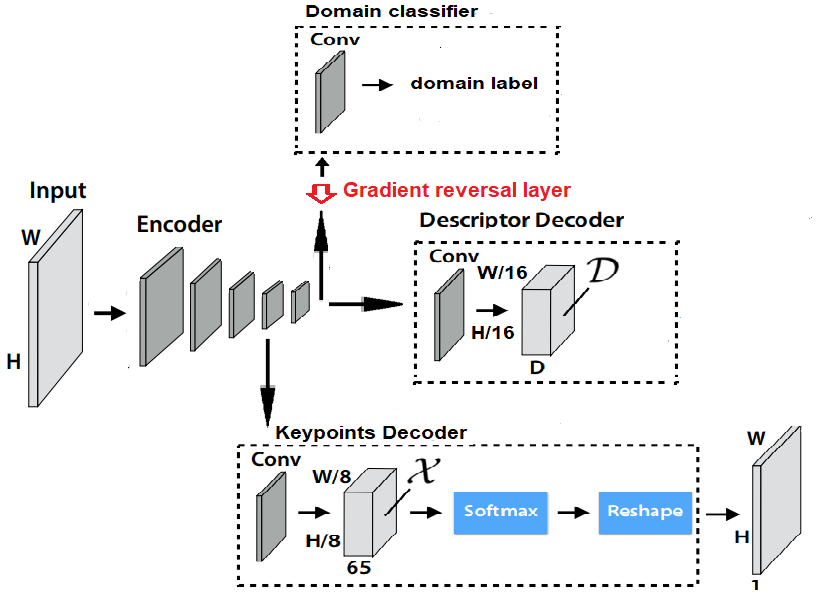}
\caption{Architecture of one twin module of the Siamese network used to train keypoint detector and domain invariant descriptor.}
\label{fig:our_superpoint_training}
\end{figure*}

The first first stage of SuperNCN, Superpoint-based keypoint detector and descriptor, is trained using a Siamese-like~\cite{chopra2005learning} approach.
In order to achieve domain-invariance, that is to make the resulting descriptors similar whenever the image of the observed scene is taken during the day or at night, the domain adaptation using an adversarial loss \cite{ganin2016domain} technique is used.
The architecture of one of the twin modules building the Siamese network used to train the keypoints descriptor is shown in Fig. \ref{fig:our_superpoint_training}.
It includes an encoder, keypoints decoder and a descriptor decoder, which together form a typical feed-forward architecture. 
Domain adaptation is achieved by adding a  domain classifier connected to the last layer of the encoder via a gradient reversal layer. Gradient reversal layer multiplies the gradient by a negative constant during the backpropagation-based training. 
Gradient reversal forces the feature map produced by the encoder to be domain agnostic (so the domain classifier cannot deduce if it's produced from an image taken during the day, at night, in summer or winter).
This is intended to make the resultant descriptors domain-invariant, so descriptors of a same scene point captured at different conditions are as close to each other as possible.

The Siamese network is trained by presenting pairs of similar (corresponding to the same scene point) and dissimilar (corresponding to different scene points) image patches.
Patches are sampled from pairs of images: an image randomly chosen from one of the Oxford RobotCar traversals (dawn, dusk, night, overcast-summer, overcast-winter, rain, snow) and it's transformed version created by applying a random perspective warp and photometric distortion.
Additionally, a numeric domain label (a traversal identifier) is given, identifying the source traversal from which the input image is sampled. 
During the training, the domain label and domain type inferred by the domain classifier are used to compute cross entropy loss. By adding the gradient reversal layer between the domain classifier and the encoder, the encoder is forced during the training to produce feature maps that are invariant to the image domain (so the domain classifier fails at inferring whether the input image is taken during the day or at night).

\subsection{Neighbourhood Consensus}
\label{sec:nc_training}
Important characteristic of Neighbourhood Consensus Network is that it can be trained with weakly supervised data. It only needs pairs of images labeled as similar and dissimilar. 
The ground truth information about corresponding locations in both images is not needed.
Loss for training Neighbourhood Consensus Network is defined as:
$$L(I^A, I^B) = -y(mean(s^A)+mean(s^B)) \text{ , where }$$
$$s^A_{ijkl} = \frac{\exp{(c_{ijkl}})}{\sum_{ab}\exp{(c_{abkl}})} \text{, } s^B_{ijkl} = \frac{\exp{(c_{ijkl}})}{\sum_{cd}\exp{(c_{cdkl}})} \text{, and}$$ 
$$y =
  \begin{cases}
    1       & \quad \text{if images depict the same location}\\
    0  & \quad \text{if images depict different location}
  \end{cases}$$
 $I^A$ and $I^B$ are images from the database with corresponding scaled down $(a, b)$ and $(c, d)$ dimensions. $c$ is a 4D similarity score space computed by neighbourhood consensus of size $(a, b, c, d)$. It is worth mentioning that $a$, $b$, $c$, and $d$ dimensions are interpreted as sizes of scaled down feature space in real implementation. Computing a full resolution 4D hypercube is not feasible with current hardware (memory-wise and computational-wise).
 
 For training, we sample pairs of similar images by taking images from the same traversal with relative translation and rotation below 5 meters and 30 degrees thresholds respectively. 
 We chose pairs of dissimilar images by taking images with relative translation or rotation above these thresholds. 
 We used all cameras: right, left and rear for sampling training data. Training is performed on 4 RTX 2080Ti GPUs for 5 epochs and using images downsampled to 512x512 pixel resolution for efficiency. 
 We found that the choice of downsampled resolution has a limited impact on the learning process significantly. Original NCNet produces good results for 400x400 pixel input images.


\section{Experimental setup}
\label{sec:experimental_setup}
We compare performance of our solution with classical and state-of-the-art image matching methods: SIFT+KNN (SIFT~\cite{SIFT1} features and matching individual features using nearest neighbour search in the descriptor space), 
AKAZE+KNN (AKAZE~\cite{AKAZE} features and nearest neighbour-based matching approach), SP+KNN (CNN-based Superpoint~\cite{Superpoint} feature detector and descriptor and nearest neighbour-based matching approach), DGC (neural network-based Dense Geometric Correspondence Network~\cite{dgcnet2018}), ORB+GMS (ORB~\cite{ORB} features and grid-based GMS~\cite{DBLP:conf/cvpr/BianLMYNC17}) matching method.

For evaluation purposes we choose two traversals from RobotCar Seasons dataset acquired during challenging environmental conditions: \emph{rain} and \emph{night}. 
We match images from these traversals with images in the reference traversal \emph{overcast-reference}, taken at day-time at favourable conditions.
See leftmost (night) and rightmost (overcast) images in Fig.~\ref{fig:RCS_examples}.

We evaluate the performance using the standard relative pose estimation task.
For each image in a non-reference traversal, we find a set of close images in the \emph{overcast-reference} traversal using the ground truth poses provided with the dataset. 
Close images are defined as images for which distance between camera centres is below 10 meters and camera viewing angle differs by less than $\pi / 4$. 
Then, we estimate relative pose between each pair of close images, one from the reference traversal and the other from the non-reference traversal, using a method under evaluation.
For methods containing a separate feature detection step (SIFT+KNN, AKAZE+KNN, SP+KNN, ORB+GMS),
in each pair of close images we detect $N$ keypoints using a keypoint detector being evaluated.  In our experiments, in each image we choose $N=1,000$ keypoints with the strongest response with exception to evaluation of ORB+GMS method, which by design requires a very large number of keypoints. For ORB+GMS we choose $N=10,000$ strongest keypoint. Then, we compute keypoint descriptors at the keypoint locations. 
For methods using KNN matching step (SIFT+KNN, AKAZE+KNN, SP+KNN) we first find putative matches between keypoints by finding the nearest neighbour in the descriptor space. Then, Lowe's ratio test with threshold set to 0.75 is applied to filter putative matches. 
DGC~\cite{dgcnet2018} method operates directly on pairs of images, producing a dense correspondence map.
Finally, we use RANSAC~\cite{Fischler:1981:RSC:358669.358692} with 5-pt Nister algorithm~\cite{nister2004efficient} to estimate essential matrix $\mathbf{E}$ from the found correspondences. 
Rotation matrix $\hat{R}$ and translation vector $\hat{T}$ are computed from the estimated essential matrix $\mathbf{E}$.

Rotation error $R_{err}$ is measured as the rotation angle needed to align ground truth rotation matrix $R$ and estimated rotation matrix $\hat{R}$.
$$
R_{err} = \cos^{-1} \frac{\mathrm{Tr}\left(\Delta R \right)-1}{2} \quad ,
$$
where $\Delta R = R^{-1} \hat{R}$ is the rotation matrix that aligns estimated rotation $\hat{R}$ with the ground truth rotation $R$ and $\mathrm{Tr} ( \Delta R )$ is a trace of $\Delta R$.
An estimation of relative pose from two images given known intrinsic parameters, is possible only up to a scale factor. 
To compute translation error, the recovered translation vector $\hat{T}$ is first brought to the same scale as the ground truth vector. 
$$
T_{err} = \left\Vert \frac{\|T\|}{\|\hat{T}\|} \hat{T} - T \right\Vert \quad ,
$$

We report the ratio of successful rotation and translation estimation attempts. For rotation, it's calculated as the percentage of rotation estimation attempts with rotation error less then a threshold: $R_{err} < \theta_R$. For translation, it's calculated as the percentage of translation estimation attempts with translation error less then a threshold times ground truth distance between the cameras: $t_{err} < \theta_T \|T\|$.

\section{Results}
\label{sec:results}

\begin{figure}
\centering
\includegraphics[width=0.45\textwidth]{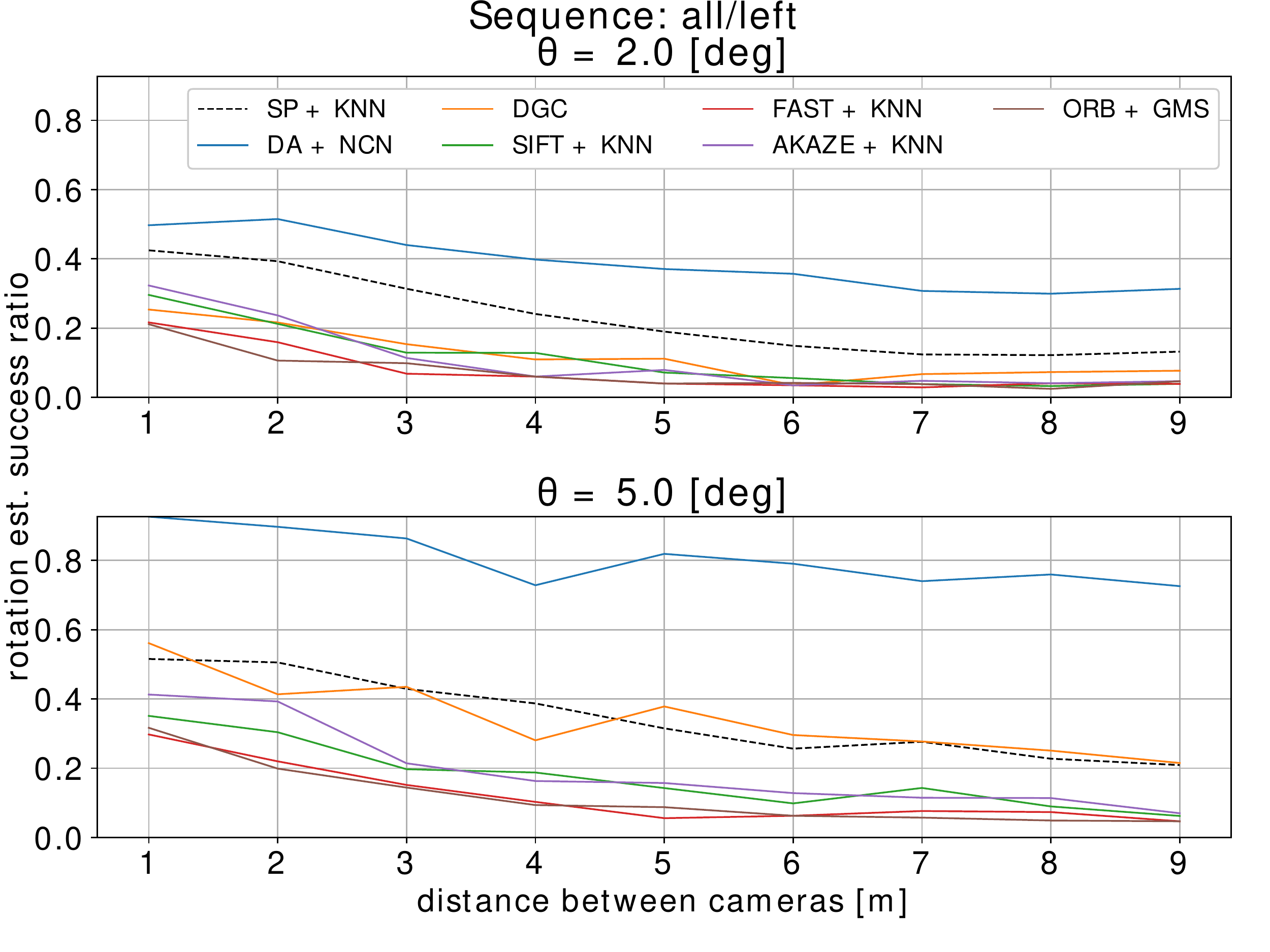}
\includegraphics[width=0.45\textwidth]{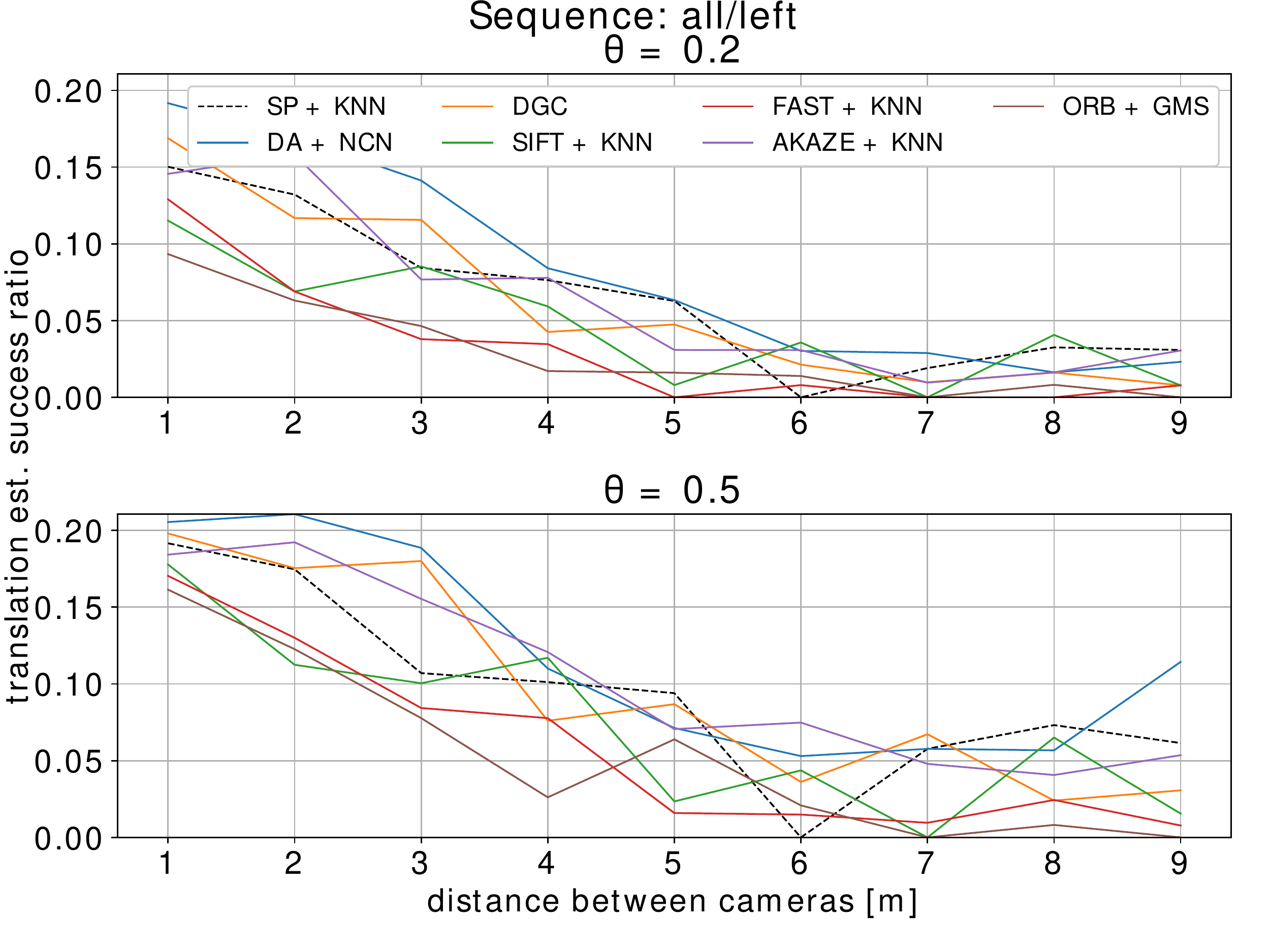}
\caption{Pose estimation results averaged on all pairs of sequences from \emph{rain} and \emph{night} traversals matched with a reference traversal.}
\label{fig:results}
\end{figure}

Evaluation results are presented in Fig.~\ref{fig:results}. 
Left plot shows a ratio of successful rotation estimation attempts and right plot shows a ratio of successful translation estimation attempts as a function of a distance between camera centres. Results are averaged for all pairs of close images, one from the reference traversal and one from non-reference traversal: \emph{rain} or \emph{night}.

Our method (denoted DA+NCN) achieves significantly higher rotation estimation success ratio than all other evaluated approaches (see left plot in Fig.~\ref{fig:results}). Despite very challenging conditions and large scene appearance changes between \emph{overcast-reference} and non-reference (\emph{night}, \emph{rain}) traversals, it is able to successfully recover relative rotation between cameras. It beats other approaches by a large margin. 
In terms of translation estimation success ratio (right plot in Fig.~\ref{fig:results}) our method is better than other compared approaches by a smaller margin.
It must be noted that translation estimation success ratio is relatively small for all evaluated methods.
For small camera center displacement (below 3 meters) it's about 20\% for our method but then it sharply drops to about 5-10\%. Other methods perform even worse.
This can be attributed to the fact, that in images from left and right camera there are a lot of planar structures such as building facades. Planar points create a degenerate configuration for essential matrix estimation task. In such cases all evaluated methods fail to produce correct results.


\section{Conclusions and future work}
\label{sec:conclusions}

The proposed method allows robust matching of local features between two images under a strong scene appearance changes.
The success of our method can be attributed to two main factors. First, domain adaptation by injecting adversarial loss during the network training produces features robust to variable environmental conditions. Traditional feature descriptors differ significantly when computed for well lit scene and at night. Using domain adaptation allows learning features more robust to variable environmental conditions.
Second, using NCNet promotes spatial smoothness of computed correspondences by boosting the matching score of spatially consistent groups of matches. This further increases performance in challenging environmental conditions, compared to traditional approaches, where features are matched in isolation.

In future work, we'd like to investigate the root cause of the performance difference in the rotation and translation estimation tasks. In rotation estimation task our method significantly outperforms all other evaluated approaches, whereas in translation estimation task it's better by a small margin.
We also plan to improve the computational performance of the proposed method. Currently it can efficiently process medium resolution images (1024x1024 pixels) but it's too slow for practical usage to process high resolution FullHD images.


\bibliographystyle{splncs04}
\bibliography{arcore}
\end{document}